\documentclass[11pt,final,onecolumn]{IEEEtran}
\IEEEoverridecommandlockouts

\usepackage{cite}
\usepackage{amsmath,amssymb,amsfonts}
\usepackage{algpseudocode,algorithm,algorithmicx}
\usepackage{graphicx}
\usepackage{textcomp,color}
\usepackage{xcolor}
\usepackage{flushend}

\newcommand{\xb}{\mathbf{x}}
\newcommand{\cb}{\mathbf{c}}

\newcommand{\gb}{\mathbf{g}}

\newcommand{\pb}{\mathbf{p}}
\newcommand{\qb}{\mathbf{q}}

\newcommand{\bb}{\mathbf{b}}
\newcommand{\Ab}{\mathbf{A}}

\newcommand{\Cb}{\mathbf{C}}
\newcommand{\Db}{\mathbf{D}}

\newcommand{\Vb}{\mathbf{V}}

\newcommand{\Qb}{\mathbf{Q}}

\newcommand{\defi} { \stackrel{\bigtriangleup}{=} }

\renewcommand{\baselinestretch}{1.3}

\begin{document}
\setcounter{page}{1}

\title{Credal Valuation Networks for Machine Reasoning Under Uncertainty}%

\author{Branko~Ristic\thanks{Correponding author, email address: branko.ristic@rmit.edu.au},
Alessio Benavoli, Sanjeev Arulampalam\thanks{B.~Ristic is  with the School
of Engineering, RMIT University,  Australia; A. Benavoli is with Trinity College Dublin, Ireland; S. Arulampalam is with Defence Science and Technology Group, Edinburgh, SA, Australia}
}
\markboth{Prepared for arXiv }{Ristic,Benavoli, Arulampalam}

\maketitle

\begin{abstract}
Contemporary undertakings provide limitless opportunities for widespread application of machine reasoning and artificial intelligence in situations characterised by uncertainty, hostility and sheer volume of data. The paper develops a valuation network as a graphical system for higher-level fusion and reasoning under uncertainty in support of the human operators.  Valuations, which are mathematical representation of (uncertain) knowledge and collected data, are expressed as credal sets, defined as coherent interval probabilities in the framework of imprecise probability theory. The basic operations with such credal sets, combination and marginalisation,  are defined to satisfy the axioms of a valuation algebra. A practical implementation of the credal valuation network is discussed   and its utility demonstrated on a small scale example.
\end{abstract}



\section{Introduction}
As the volume of information (domain knowledge and data) exceeds, in most practical situations, the ability of human operators to process and comprehend it in  a timely manner, we increasingly rely on machine intelligence for reasoning and forming inferences.  Information can appear in different forms, for example, as the numerical  measurements from physical sensors, in the form of natural language statements (written or spoken) or  as the contextual prior information in the form of maps or images.  All types of information, however, have one characteristic in common: they are affected by a certain degree of uncertainty. Two types of uncertainty are typically distinguished \cite{aleatoric_21}: {\em aleatoric}  uncertainty, which is due to stochastic variability, and {\em epistemic} uncertainty, caused by the lack of knowledge.

Probability theory was developed for quantitative modeling and statistical inference in the presence of aleatoric uncertainty.  In the probabilistic framework,  stochastic variability is modelled using probability functions.  In applications where such probabilistic models are only partially known, for example, due to the scarcity of training data,  epistemic uncertainty  must also be taken into account. This fact gave rise to alternatives  to classical
probability for quantitative modeling of uncertainty. They are collectively referred to as
{\em non-additive} probabilities \cite{shafer2008non,hampel2009nonadditive}, because they do not satisfy sigma-additivity. They include for example, coherent lower (or upper) previsions, used in  imprecise probability theory  \cite{walley1991statistical,Augustin2014},  belief functions, used in  Dempster-Shafer (a.k.a.  belief function) theory \cite{shafer_76,Smets1994},  and possibility functions, used in  possibility theory \cite{dubois2015possibility},  A review and comparison of aforementioned non-additive probability frameworks is presented in \cite{ristic2020tutorial}.

Historically, the first machines for reasoning  captured the knowledge of human experts by a complex system of ''if-then'' rules \cite[Ch.9]{waltz1990multisensor}. Their main drawback was the lack of a means  in handling
{\em uncertainty}.
The invention of {\em Bayesian networks} (BN) \cite{pearl} in the mid
1980s, for knowledge representation and probabilistic inference,
represented an important step in the development
of expert systems capable of reasoning under uncertainty. In the BN context, several architectures \cite{lepar98comparison} have been proposed
 for exact computation of marginals of multivariate discrete probability distributions.
 One of the pioneering architectures for computing
 marginals was proposed by Pearl \cite{pearl} for multiply
 connected Bayesian networks. In 1988, Lauritzen and Spiegelhalter \cite{laur_88} proposed an  alternative architecture for computing marginals of the multidimensional probability density by so-called ''local computation'' in join trees.
 This architecture has been generalized by Lauritzen
 and Jensen \cite{laur_96} so that it applies more generally
 to other uncertainty representation frameworks, including the Dempster-Shafer's belief
 function theory \cite{shafer_76}.
 Inspired by the work of Pearl, Shenoy and Shafer \cite{shafer_87}
 first adapted and generalized Pearl's architecture to the case
 of finding marginals of joint Dempster-Shafer belief functions
 in join trees. Later, motivated by the work of Lauritzen
 and Spiegelhalter \cite{laur_88} for the case of probabilistic
 reasoning, Shenoy and Shafer proposed the valuation based system (VBS) for computing
 marginals in join trees and established the set of axioms that
 combination and marginalisation (focusing) operations need to satisfy in order to make the local computation concept applicable
  \cite{shenoy_shafer_90}. Reasoning networks based on the Shenoy-Shafer architecture are referred to as valuation networks.
  A slightly modified version of the Shenoy-Shafer axiomatic formulation was developed by Kohlas \cite{Kohlas_03} with the resulting mathematical structure referred to as the {\em valuation algebra}.
  The central component of a valuation algebra is a {\em valuation}: a quantified representation of uncertain piece of information in the adopted framework of uncertainty modeling.
 The axioms of valuation algebra are
 satisfied in the framework of  probability theory, possibility theory and Dempster-Shafer theory \cite{Kohlas_03},
 leading to development and application of the corresponding valuation networks \cite{Shenoy1992,almond_95,Haenni2004,Benavoli2009}.


 This paper develops a valuation network for reasoning under uncertainty where valuations are expressed as a special case of coherent lower (upper) previsions, that is, as credal sets defined by the coherent probability intervals on singletons \cite{de1994probability}.  This representation of uncertain information is convenient because it requires only twice the number of values required to represent a standard probability function (as opposed to belief functions or generic coherent lower previsions, where this number grows exponentially). Probability intervals have been used
for example in Bayesian networks with imprecise probabilities \cite{corani2012bayesian}, and for classification with imprecise probabilities \cite{class_ip_18}.  The basic operations with coherent probability intervals, i.e. the combination rule and marginalisation,  will be defined in the paper as a generalisation of the standard probabilistic approach. The set of coherent probability intervals, with such basic  operations, will be shown to satisfy the axioms of valuation algebra. Subsequently the resulting valuation network, referred to as the {\em credal} valuation network,  will be implemented using the Shenoy-Shafer architecture and its performance demonstrated and compared to the evidential network \cite{Benavoli2009} on a small scale example taken from \cite{almond_95}. Our work is somewhat related to   \cite{maua2012updating} and \cite{casanova2021information}. While both references define
 valuation algebras of coherent lower previsions and credal sets, respectively, valuations and basic operations are different from those presented here.

\section{Valuation algebra}

This section reviews the fundamental concepts of valuation algebra, following \cite{Kohlas_03,benavoli_ristic_2013}.

\subsection{Valuations and basic operations}
\label{s:valuations}
Realistic applications of systems for reasoning under uncertainty typically involve many interacting variables, connected in a network which
codifies the relationships between them. Let
$\Vb$ be the set of all variables\footnote{Sets of variables  are denoted with capital boldface letters.}  in this network. A valuation $\varphi$ represents a piece of information (available knowledge or measurements) about the relationship among a subset of variables $d(\varphi)\subseteq \Vb$, where $d(\varphi)$ is referred to as the domain of $\varphi$. Let ${\Phi}$ denote the set of all valuations in a network. Then,  $d:\Phi \rightarrow 2^{\Vb}$, where $2^{\Vb}$ is the power set of $\Vb$, is referred to as the {\em labeling} operation.

The relationship among the variables in the set $\Db=d(\varphi)$ is specified by assigning values (corresponding to beliefs)  to the elements of a set of possible configurations of $\Db$, referred to as the state space or the frame of $\Db$. Suppose the frame of variable $X\in \Db$ is $\Theta_X$. Then, the frame of $\Db$ is defined as $\Theta_\Db  \defi \times \{\Theta_X : X \in   \Db \}$, where $\times$ denotes the Cartesian product.

Let us next introduce an example of a valuation network \cite{almond_95}, which will be solved in Sec.  \ref{s:demo}.

{\em Example (Arrival delay).} The problem is  to  estimate  the  arrival  delay  of  a  ship  carrying  a valuable cargo. The following pieces of (prior) information are expressed by valuations:\\
{$\varphi_1$:} Arrival delay ($A$) is due to departure delay ($D$) and the travel delay ($T$);\\
{$\varphi_2$:} Departure delay ($D$) is caused by unexpected difficulties in loading ($L$) the cargo,  or by the engine service ($S$);\\
{$\varphi_3$:}  Travel delay ($T$) is due to bad weather ($W$) or unplanned repairs ($R$) on the sea;\\
{$\varphi_4$:} A repair on sea ($R$) is related to the service ($S$).\\
Before the departure, the following additional (uncertain) information becomes available:\\
{$\varphi_5:$} Rumours about the loading delay $D$;\\
{$\varphi_6:$} Captain's decision on the type of service $S$ (e.g. comprehensive, basic or nil);\\
{$\varphi_7:$} Weather  $W$ forecast for the entire trip.  $\square$

\begin{figure}[htb]
\centerline{\includegraphics[width=0.5\linewidth]{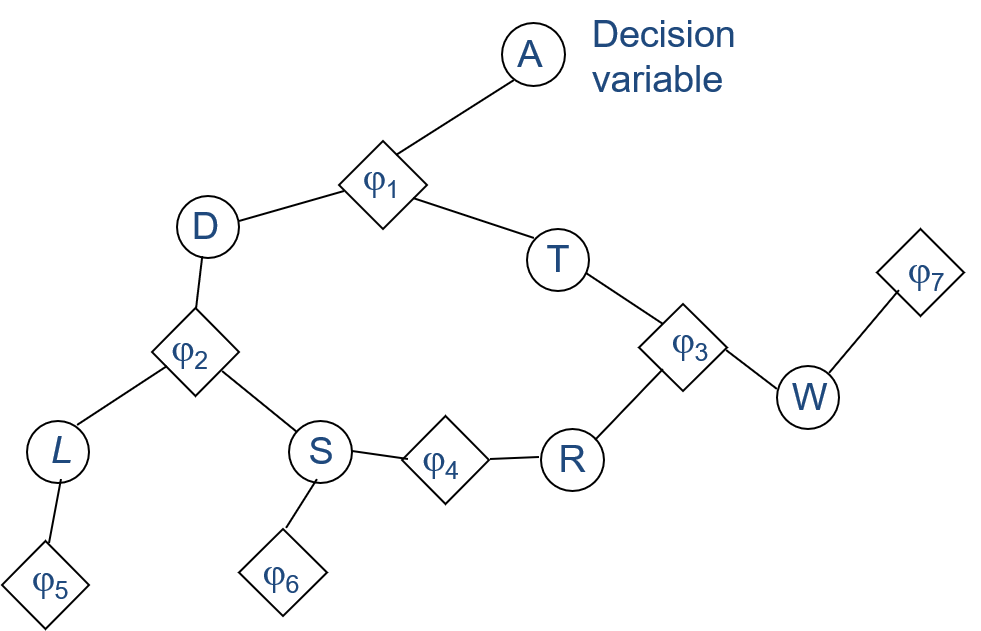}}
 \caption{\footnotesize A graphical representation of the valuation network corresponding to the ''Arrival delay'' example.}
 \label{f:va}
\end{figure}

The set of valuations in this example is ${\Phi}= \{\varphi_1,\varphi_2,\cdots,\varphi_7\}$; the set of variables is $\Vb = \{A,~D,~T,~L,~S,~W,~R\}$. A graphical representation of the valuation network corresponding to this example is shown in Fig. \ref{f:va}. Variables are represented by circles, whereas valuations by diamonds. Each valuation is
connected by edges to the subset of variables which define its
domain. For example, the domain of valuation  $\varphi_1$
is $d(\varphi_1) = \{A,D,T\}$. Because we are interested in the arrival delay, variable $A$ is referred to as the decision (or inference) variable.

There are two basic operations with valuations.
\begin{itemize}
 \item {\bf Combination} is a binary operation  $\otimes: {\Phi}  \times {\Phi}  \rightarrow {\Phi}$. If $\varphi_1,\varphi_2 \in {\Phi}$ are two valuations, then the combined
valuation $\varphi_1 \otimes \varphi_2$ represents the aggregated knowledge from $\varphi_1$ and $\varphi_2$.
\item {\bf Marginalization} is a binary operation $\downarrow: {\Phi}  \times 2^{\Vb} \rightarrow {\Phi}$ which
 is focusing the knowledge to a smaller domain. For example,
if $\varphi \in {\Phi}$  and $\Cb \subseteq d(\varphi)$, then the marginalized valuation $\varphi^{\downarrow \Cb}$
represents the knowledge obtained by focusing $\varphi$ from $d(\varphi)$ to $\Cb$.
 \end{itemize}
Instead of marginalization, we can use another basic operation called
\textit{variable elimination}, defined as:
 $\varphi^{-X} \defi \varphi^{\downarrow d(\varphi)\backslash \left\{X\right\}}$, where $X \in {\Vb}$ and symbol $\setminus$ denotes the set difference.
 Note that $X \notin d(\varphi)$ implies $\varphi^{-X}=\varphi$.

\subsection{Axioms of valuation algebra}
\label{sec:axioms}
Given a finite collection ${\Phi}    = \{\varphi_1,\dots, \varphi_r\}  $ of valuations, inference  refers to
marginalization of the joint valuation $\otimes {\Phi} =\varphi_1\otimes\dots \otimes \varphi_r $ to a subset of variables $\Db^o \subseteq \Vb$ called {\em decision variables}. In the''Arrival delay'' example,  $\Db^o =\{A\}$.

The straightforward approach to inference would be to compute the
joint valuation first and then to marginalize it to $\Db^o$. Unfortunately, this would be cumbersome in practice even for a small scale valuation network because
the domain size increases with each combination, whereas the complexity grows exponentially with the domain size.  For instance, if there
are $n$ variables and each variable can assume $m$ different
values (configurations), then there are $m^n$ configurations in the joint domain of all
variables.

By imposing certain axioms for the operations of
\textit{labeling}, \textit{combination}, and
\textit{marginalization}
\cite{shenoy_shafer_90,shenoy_92,haenni_04}, it is
possible to compute the marginal $(\otimes {\Phi})^{\downarrow \Db^o}$
on local domains, without the need to explicitly compute the joint
valuation.  The list of axioms is as follows \cite{Kohlas_03}:
\begin{description}
 \item[(A1)] {$({\Phi}, \otimes)$ is a commutative monoid,} i.e. it is closed, associative and commutative under combination $\otimes$. Furthermore, for a set $\Phi_\Db$ which represents the set of all valuations with domain  $\Db \subseteq \Vb$, exists an identity valuation $e_\Db\in\Phi_\Db$ such that $e_\Db \otimes \varphi =\varphi \otimes e_\Db = \varphi$ for all $\varphi \in {\Phi}_\Db$.
 \item[(A2)] \textit{Labeling:}  if  $\varphi_1, \varphi_2 \in \Phi$ ,  then  $d(\varphi_1 \otimes \varphi_2)=d(\varphi_1) \cup d(\varphi_2)$.
\item[(A3)] \textit{Marginalization:} If $\varphi \in  {\Phi}$ and $\Cb \subseteq d(\varphi)$, then $d(\varphi^{\downarrow \Cb})=\Cb$.
\item[(A4)] \textit{Transitivity of marginalization:} If $\varphi \in  {\Phi}$  and $\Db_1 \subseteq \Db_2 \subseteq d(\varphi)$, then
$\varphi^{\downarrow \Db_1}=(\varphi^{\downarrow \Db_2})^{\downarrow \Db_1}$.
\item[(A5)] \textit{Distributivity of marginalization over combination:}  If $\varphi_1,\varphi_2 \in {\Phi}$, with domains $\Db_1=d(\varphi_1)$ and $\Db_2 =d(\varphi)$,  then $(\varphi_1 \otimes \varphi_2)^{\downarrow \Db_1}=\varphi_1 \otimes \varphi_2^{\downarrow \Db_1 \cap \Db_2}$.
\item[(A6)] \textit{Identity:}   For $\Db_1,\Db_2 \subseteq \Vb$, we have $e_{\Db_1} \otimes e_{\Db_2} = e_{\Db_1 \cup \Db_2}$.
 \end{description}
A system $\{{\Vb}, {\Phi}, d, \otimes, \downarrow \}$  is called \textit{Valuation Algebra}
(VA) if the operations of labeling $d$, combination $ \otimes$, and
marginalization $\downarrow$ satisfy the above  axioms
\cite{Kohlas_03}.

We can replace the operation of marginalisation with the variable elimination. Then axioms (A4) and (A5) will be replaced with:
\begin{description}
 \item[(A4')]\textit{Commutativity of elimination:} if $\varphi \in  {\Phi}$ and $X,Y \in \Vb$, then
$(\phi^{-X})^{-Y}=(\phi^{-Y})^{-X}$.
 \item[(A5')]\textit{Distributivity of elimination over combination:} If $\varphi_1,\varphi_2 \in  {\Phi}$
 with $ X \notin d(\varphi_1)$, then  $(\varphi_1 \otimes \varphi_2)^{-X}=\varphi_1 \otimes \varphi_2^{ -X}$.
 \end{description}
Because of (A4'), it possible to write
$\varphi^{ -\Db}$
for the elimination of several variables $\Db \subset \Vb$, since the
result is independent of the order of elimination. As a consequence,
marginalization can be expressed in terms of variable eliminations
by $\varphi^{- \Db}=\varphi^{\downarrow d(\varphi) \backslash \Db}$.
Therefore,
operations of marginalization and variable elimination together with
their respective systems of axioms are equivalent.

The concept of VA is very general and has a wide range of
instantiations, such as the VA of probability mass functions, VA of  systems of linear equations, VA of linear  inequalities, VA of Dempster-Shafer belief functions, VA of Spohns disbelief functions, VA of possibility functions, and others \cite{Kohlas_03}. The VA of probability mass functions (PMFs)
is briefly reviewed next.

\subsection{Valuation algebra of probability mass functions}
\label{s:VA_PMF}

Consider $\Db\subseteq \Vb$ and its frame $\Theta_\Db$.  Let the probability of an event $A \subseteq\Theta_\Db$ be denoted $P(A)$.
The probability mass function (PMF) $p:\Theta_\Db \rightarrow [0,1]$, corresponding to the probability measure $P$ is introduced via the relationship $P(A)=\sum_{x\in A} p(x)$. The PMF $p$ assigns to each configuration $x\in\Theta_\Db$ the probability $p(x)$ that $x$ is the true value.

 Suppose two valuations are expressed by two PMFs on $\Theta_\Db$, and denoted $p_1$ and $p_2$. Assuming that they  specify the beliefs from two independent sources, the combination operator is given by \cite{distributedClass,benavoli_ristic_2013}:
\begin{equation}
	(p_1 \odot p_2)(x)=  \frac{p_1(x)p_2(x)}{\sum\limits_{y\in\Theta_\Db} p_1(y)p_2(y)}
	\label{e:comb_pmf}
\end{equation}
for any configuration $x\in\Theta_D$, providing that the denominator  $\sum_{y\in\Theta_\Db} p_1(y)p_2(y)>0$. If this condition is not satisfied, than $p_1$ and $p_2$ are in a total conflict and cannot be combined.

The concept of marginal distribution is well known in probability theory, and so is the marginalisation operator. Let $p^\Db$ denote a PMF defined on domain $\Db$. Then its marginalisation to the domain $\Cb \subset \Db$ is defined as \cite{benavoli_ristic_2013}
\begin{equation}
    p^{\Db \downarrow \Cb} = \sum_{y:y \downarrow x} p^\Db(y),
\end{equation}
where the summation is over all configurations $y\in\Theta_\Db$ such that $y$ reduces to configuration $x\in\Theta_\Cb$ by elimination of variables $\Db \setminus \Cb$.

A set of PMFs with operations of combination and marginalisation satisfies axioms (A1)-(A6) and hence is a valuation algebra \cite{Kohlas_03}.   For example, the combination operator (\ref{e:comb_pmf}) can be easily shown to be associative and commutative, because these two laws hold for multiplication and summation of numbers. The neutral element is the uniform PMF on $\Theta_\Db$.

\section{Valuation algebra of credal sets}

Imprecise probabilities provide a general framework for modelling uncertain knowledge. Within this framework, different formalisms  for modelling with imprecise probabilities have been proposed: a coherent set of desirable gambles, coherent lower previsions, and credal sets \cite{Augustin2014}. All three formalisms are
mathematically equivalent.

\subsection{The set of valuations}
\label{s:val}
We adopt as  valuations a special class of credal sets, defined by probability intervals on singletons \cite{de1994probability, antonucci14}.
A credal set is a closed convex set of PMFs of a discrete variable $X$. We start from a premise that the valuation algebra of credal sets should represent a generalisation of the valuation algebra of PMFs, discussed briefly in Sec. \ref{s:VA_PMF}. In the case  the credal set contains only one element (a single PMF), then the two valuation algebras should be identical.

A credal set can be geometrically represented as a  convex polytope on the probability simplex\footnote{A polytope  is a geometric object with ``flat'' sides. For example, a two-dimensional polytope is a polygon. A probability simplex is the space in which each point represents a probability distribution.}. Any convex polytope can be specified either as (i) the intersection of half-spaces (expressed by a system of linear inequalities), or (ii) as the convex hull of its vertices or extreme points. We will elaborate this  later by an example.

Consider a random variable  $X\in \Vb$ of a valuation network; its frame is $\Theta_X$. The totally uninformative credal set on $\Theta_X$, referred to as the {\em vacuous} credal set, contains all PMFs on $\Theta_X$ and is defined as:
\begin{equation}
    \mathcal{P}^X = \{p:p(x)\geq 0, \forall x\in\Theta_X, \text{ and }\sum_{x\in\Theta_X} p(x) = 1\}.
    \label{e:vac}
\end{equation}
Any other (more informative) credal set over $\Theta_X$ is defined by imposing additional constraints to $\mathcal{P}^X$. The most informative credal set is the one that contains a single (precise) PMF. The case where all valuations in the  network are precise is treated as the valuation algebra of PMFs, discussed in Sec. \ref{s:VA_PMF}.

\begin{figure}[tbh]
\centerline{\includegraphics[height=7cm]{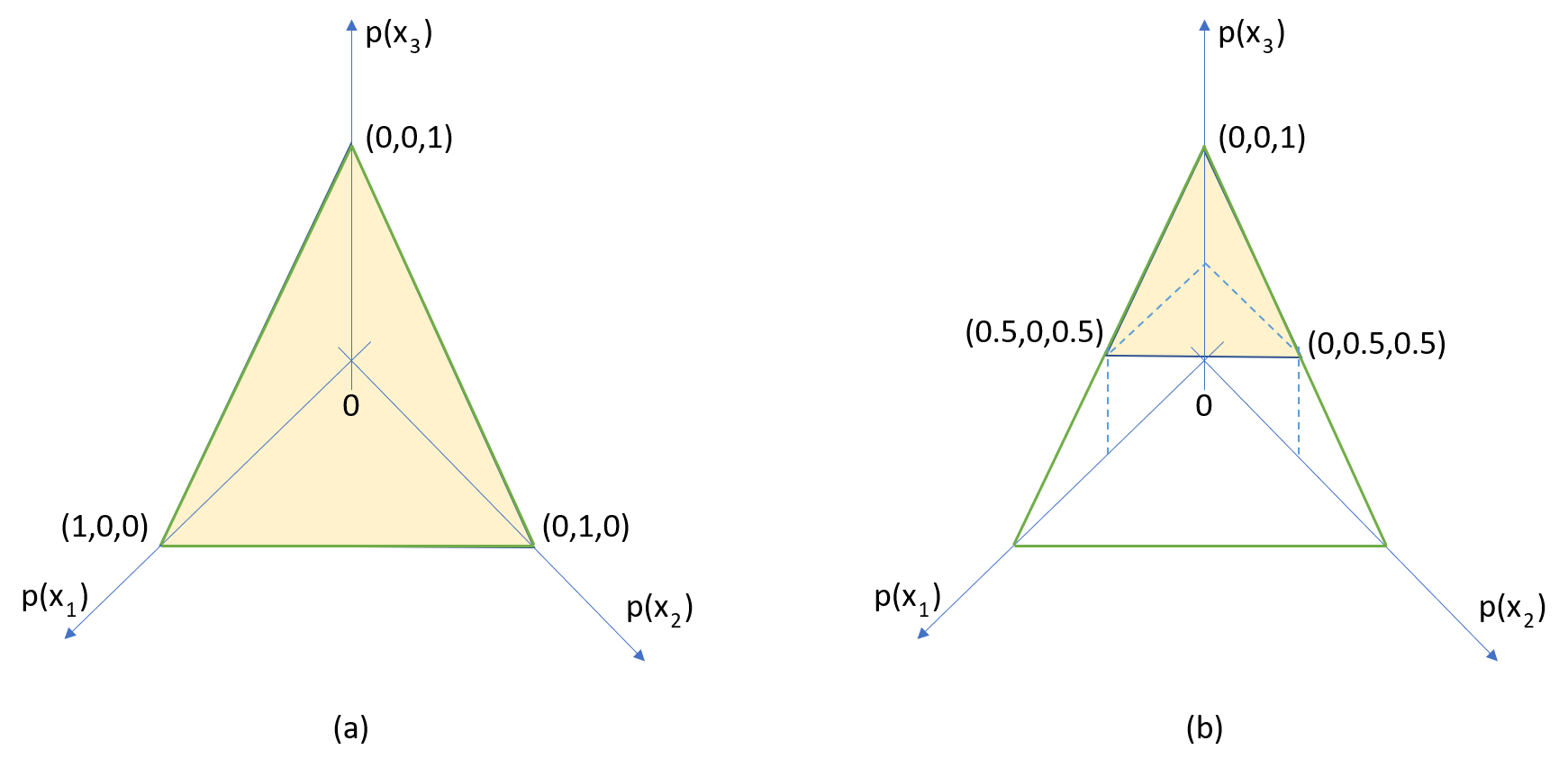}}
\caption{\footnotesize Geometrical representation of (a) the vacuous credal set $\mathcal{P}^{X}$ over $\Theta_X=\{x_1,x_2,x_3\}$; (b) credal set $L^{X}$, specified by (\ref{e:K1}).}
 \label{f:credal}
\end{figure}

{\em Example 1.}  Consider a random variable $X$ defined on a three-dimensional frame $\Theta_X = \{x_1,x_2,x_3\}$. Let the credal set be defined as:
\begin{equation} L^X = \{p\in\mathcal{P}^X:p(x_1) + p(x_2) \leq p(x_3)\}. \label{e:K1}
\end{equation}
First we show how credal set $L^X$ can be expressed as the intersection of half-spaces. Note that half-spaces which define  $\mathcal{P}^X$ on $\Theta_X = \{x_1,x_2,x_3\}$ can be represented with the following system of linear inequalities:
\begin{equation}
\begin{matrix}
-p(x_1) &  &  & \leq &0\\
      & -p(x_2)  &   & \leq & 0 \\
       &   & -p(x_3)  & \leq & 0 \\
     +p(x_1) & + p(x_2) & +p(x_3) & \leq & 1\\
     -p(x_1) & - p(x_2) & -p(x_3) & \leq & -1
\end{matrix}
\label{e:ex1a}
\end{equation}
The first three inequalities in (\ref{e:ex1a}) follow from the first condition in (\ref{e:vac}), that is $p(x_i)\geq 0$, for $i=1,2,3$.  The last two inequalities in (\ref{e:ex1a}) simply express the normalisation condition, i.e.  $p(x_1)+p(x_2)+p(x_3)=1$. Finally, the last condition which defines $L^X$ in (\ref{e:K1}) can be represented with inequality:
\begin{equation}
\begin{matrix}
p(x_1) & + p(x_2) & -p(x_3)  & \leq & 0.
\end{matrix}
\label{e:ex1b}
\end{equation}
The specification of any credal set as the intersection of half-spaces can always be expressed compactly in a  matrix form as $    \Ab \pb \leq \bb$.
For $L^X$ of (\ref{e:K1}), according to (\ref{e:ex1a}) and (\ref{e:ex1b}), we have
\begin{equation}
    \Ab = \left [ \begin{matrix} -1 & 0 & 0\\ 0 & -1 & 0\\ 0 & 0 & -1 \\1 & 1 & 1\\ -1 & -1 & -1 \\ 1 & 1 & -1  \end{matrix} \right], \hspace{.5cm} \pb = \left[ \begin{matrix} p(x_1) \\p(x_2)\\p(x_3)\end{matrix} \right], \hspace{0.5 cm} \bb = \left[\begin{matrix} 0\\0\\0\\1\\-1\\0 \end{matrix}\right].
\end{equation}
A credal set can also be specified by its extreme points. For $L^X$ of (\ref{e:K1}), there are three such points, given by vectors: $\pb' = [0,\; 0,\; 1]^T$, $\pb'' = [0,\; 0.5,\; 0.5]^T$ and $\pb''' = [0.5,\; 0,\; 0.5]^T$. Fig. \ref{f:credal} provides a graphical representation of credal set  $L^X$ of (\ref{e:K1}) and its corresponding vacuous credal set $\mathcal{P}^X$.
$\Box$

We will adopt as valuations a special class of credal sets which are defined as  probability intervals on singletons, i.e.
\begin{equation}
    K^X = \{p\in\mathcal{P}^X:\underline{p}_x \leq p(x)\leq \overline{p}_x, \forall x \in \Theta_X\}.
\label{e:cs2}
\end{equation}
Note that the credal set $L^X$ of (\ref{e:K1}) in Example 1 can be specified in form of (\ref{e:cs2}) as follows:
\begin{equation} L^X =  \left\{p \in \mathcal{P}^X:p(x_1)\in\left[0,\,0.5\right],\; p(x_2)\in\left[0,\,0.5\right],\; p(x_3)\in\left[0.5,\,1\right]  \right\}.
\label{e:exK3}
\end{equation}
For the vacuous credal set, $\underline{p}_x=0$ and $\overline{p}_x=1$, for all $x\in \Theta_X$.  In general, however, the space of credal sets defined by the intersection of  half-spaces subsumes the set defined by (\ref{e:cs2}).

The choice of probability intervals on singletons in (\ref{e:cs2}) is not arbitrary.  First, in order to avoid that  $K^X$ defined by  (\ref{e:cs2}) is empty, we have the following condition \cite{de1994probability} (see Appendix A):
\begin{equation}
    \sum_{x\in\Theta_D}\underline{p}_x \leq 1 \leq \sum_{x\in\Theta_D} \overline{p}_x.  \label{e:cond1}
\end{equation}
Furthermore, probability intervals should also satisfy  the conditions of  {\em reachability}  \cite{de1994probability}. Let the credal set be defined with probability intervals  $[\underline{p}_i,\overline{p}_i]$, for $i=1,\dots,|\Theta_X|$. Then the following must hold
\begin{equation}
    \sum\limits_{j \neq i} \underline{p}_j + \overline{p}_i \leq 1, \text{ and }
 \sum\limits_{j\neq i} \overline{p}_j + \underline{p}_i \geq 1,   \label{e:cond2}
\end{equation}
for $i=1,\dots,|\Theta_X|$. Condition on the left of (\ref{e:cond2}) is equivalent to
stating that for each $i=1,\dots,|\Theta_X|$ there exist a PMF $p^i\in K^X$ which  {\em reaches} the  upper probability $\overline{p}_i$, i.e. $p^i(x_i)=\overline{p}_i$.
Analogously, condition on the right of (\ref{e:cond2}) is equivalent to
stating that for each $i=1,\dots,|\Theta_X|$ there exist a PMF $q^i\in K^X$ which  {\em reaches} the lower probability $\underline{p}_i$, i.e. $q^i(x_i)=\underline{p}_i$ (see Appendix A).
According to Walley  \cite[Sec.2.7]{walley1991statistical}, probability intervals which satisfy (\ref{e:cond1}) and (\ref{e:cond2}) are {\em coherent}.
We will only consider credal sets defined by (\ref{e:cs2}), with probability intervals  that satisfy (\ref{e:cond1}) and (\ref{e:cond2}). It is easy to verify that $L^X$, considered in Example 1, is such a credal set.

\subsection{Basic operations }

The set of valuations $\Phi$ was specified in Sec. \ref{s:val} as the set of credal sets defined by coherent probability intervals on singletons. We will refer to this class of valuations, in short,  as credal sets. They represent an epistemic generalisation of the valuations specified as PMFs in Sec. \ref{s:VA_PMF} and next we define the combination and marginalisation operators for them.

\subsubsection{Combination operator}

Suppose two beliefs from independent sources are expressed on domain $\Db$ as credal sets $K_1^\Db \in\Phi_{\Db} $ and $K_2^\Db\in\Phi_{\Db}$. The credal set of the combined (fused) belief on $\Db$, i.e.
\begin{equation}
    K_{12}^\Db =  K_1^\Db\otimes K_2^\Db,  \label{e:comb_op}
\end{equation}
can be expressed in the from (\ref{e:cs2}):
\begin{equation}
    K_{12}^\Db = \{p\in\mathcal{P}^\Db:\underline{p}_i \leq p(x_i)\leq \overline{p}_i(x_i), \text{ for } i=1,\dots,|\Theta_\Db|\},
\label{e:cs3}
\end{equation}
where
the lower probability of configuration $x_i\in\Theta_\Db$ is defined as:
\begin{eqnarray}
	\underline{p}_i & = & \min_{\substack{p_1\in K_1^\Db;\; p_2\in K_2^\Db\\
\text{s.t.}\;\;\; \sum_{x_j\in\Theta_\Db} p_1(x_j)p_2(x_j)>0}} \;\;\; p_1(x_i) \odot p_2(x_i) \\
& = & \min_{\substack{p_1\in K_1^\Db;\; p_2\in K_2^\Db\\
\text{s.t.}\;\;\; \sum_{x_j\in\Theta_\Db} p_1(x_j)p_2(x_j)>0}} \;\;\;
\frac{ p_1(x_i)p_2(x_i)}{\sum\limits_{x_j\in\Theta_\Db} p_1(x_j)p_2(x_j)}.
	\label{e:combP_lo}
\end{eqnarray}
Similarly, the upper probability of $x_i\in\Theta_\Db$ is:
\begin{eqnarray} \overline{p}_i &= & \max_{\substack{p_1\in K_1^\Db;\; p_2\in K_2^\Db\\
\text{s.t.}\;\;\; \sum_{x_j\in\Theta_\Db} p_1(x_j)p_2(x_j)>0}} \;\;\;p_1(x_i) \odot p_2(x_i) \\
& = & \max_{\substack{p_1\in K_1^\Db;\; p_2\in K_2^\Db\\
\text{s.t.}\;\;\; \sum_{x_j\in\Theta_\Db} p_1(x_j)p_2(x_j)>0}} \;\;\; \frac{ p_1(x_i)p_2(x_i)}{\sum\limits_{x_j\in\Theta_\Db} p_1(x_j)p_2(x_j) }\label{e:combP_up1} \\
& = & 1 - \min_{\substack{p_1\in K_1^\Db;\; p_2\in K_2^\Db\\
\text{s.t.}\;\;\; \sum_{x_j\in\Theta_\Db} p_1(x_j)p_2(x_j)>0}} \frac{ \sum\limits_{x_\ell \in\Theta_\Db\setminus \{x_i\}} p_1(x_\ell)p_2(x_\ell)}{\sum\limits_{x_j\in\Theta_\Db} p_1(x_j)p_2(x_j)}.
\label{e:combP_up}
\end{eqnarray}
Eq. (\ref{e:combP_lo}) minimises the probability defined by (\ref{e:comb_pmf}) over all $p_1\in K_1^\Db$ and  $p_2\in K_2^\Db$, such that $p_1$ and $p_2$ are not in total conflict. Eq. (\ref{e:combP_up1}) performs maximisation of probability (\ref{e:comb_pmf}) with the same condition on $p_1$ and $p_2$. Eq. (\ref{e:combP_up}) follows from (\ref{e:combP_up1}) using two identities: first,  any $p(x)$  is equivalent to $1-\sum_{y\in\Theta_\Db\setminus \{x\}}p(y)$, and second, $\max(1-g)= 1-\min g$.

A few remarks are in order here. First,  it is easy to verify that if credal sets $K_1$ and $K_2$ are singletons (i.e. two PMFs), then both (\ref{e:combP_lo}) and (\ref{e:combP_up1}) reduce to (\ref{e:comb_pmf}). Second, note that by construction, the lower and upper probabilities of the combined credal set are reachable and hence will satisfy coherence, i.e. conditions (\ref{e:cond1}) and (\ref{e:cond2}). Finally, we  explain why we dismiss the conjunctive and disjunctive combination operators, proposed in \cite{de1994probability}: (i) the conjunctive operator does not always exist, (ii) the result of the disjunctive operator is not necessarily an element of  $\Phi_{\Db}$, and (iii) both  operators  are incompatible with the combination rule (\ref{e:comb_pmf}).

Next we explain how to combine two valuations on different domains. Let $K_1^{\Db_1} \in\Phi_{\Db_1}$ and $K_2^{\Db_2} \in \Phi_{\Db_2}$, and $\Db_1 \neq \Db_2$.   Before we apply the combination operator (\ref{e:comb_op}), we must extend both valuations $K_1^{\Db_1}$ and $K_2^{\Db_2}$ to the joint domain $\Db_1\cup \Db_2$ in such a way that they express the same information before and after this extension. This operation, referred to as the {\em vacuous extension},  is denoted by $\uparrow$. It spreads uniformly the probability mass $p^\Cb(x)$ assigned to $x\in\Theta_\Cb$ to all configurations $y\in\Theta_\Db$ obtained from $x\in\Theta_\Cb$ by adding variables $\Db \setminus \Cb$.  Thus, the vacuous extension of a credal set  $K^{\Cb}\in \Phi_\Cb$, to domain $\Db \supseteq \Cb$, is defined as:
 \begin{equation}
      K^{\Cb \uparrow \Db} = \{p^\Db:\; p^\Db(y) = p^\Cb(x)\,\frac{|\Theta_\Cb| }{ |\Theta_\Db|}; \; \forall p^\Cb \in K^\Cb \}.
      \label{e:vect}
  \end{equation}
Note that  $\Cb\subseteq \Db$ implies $|\Theta_\Cb| \leq |\Theta_\Db$. Assuming the credal set $K^\Cb$ is specified with probability intervals $[\underline{p}^\Cb_x,\overline{p}^\Cb_x]$ for every $x\in\Theta_\Cb$, the vacuous extension  $K^{\Cb \uparrow \Db}$  will also be expressed with probability intervals, $[\underline{p}^{\Cb\uparrow \Db},\overline{p}^{\Cb\uparrow \Db}]$ for every $y\in\Theta_\Db$, where
the lower and upper limits are given by
 \begin{equation}
 \underline{p}^{\Cb\uparrow \Db}_y  =  \underline{p}^\Cb_x \frac{|\Theta_\Cb|}{|\Theta_\Db|}, \text { and } \overline{p}^{\Cb\uparrow \Db}_y =\overline{p}^\Cb_x \frac{|\Theta_\Cb|}{|\Theta_\Db|},
  \end{equation}
respectively.

\subsubsection{Marginalisation operator}

Let  $K^\Db \in \Phi_\Db$ be defined with probability intervals $[\underline{p}_y^\Db,\overline{p}_y^\Db]$, for all configurations $y\in\Theta_\Db$. Its marginalisation to domain $\Cb\subseteq \Db$ is defined as:
  \begin{equation}
      K^{\Db \downarrow \Cb} = \{p^{\Db\downarrow\Cb}:\; p^{\Db\downarrow \Cb}(x) = \sum_{y:y\downarrow x} p^\Db(y); \;\forall x\in\Theta_\Cb \}
      \label{e:marg}
  \end{equation}
  where the summation in (\ref{e:marg}) is over all $y\in\Theta_\Db$ such that configurations $y$ reduce to configurations $x\in\Theta_\Cb$ by
  elimination of variables $\Db \setminus \Cb$. The resulting valuation  $K^{\Db \downarrow \Cb}$ can be expressed with probability intervals $\left[\underline{p}^{\Db\downarrow\Cb}_x, \overline{p}^{\Db\downarrow\Cb}_x\right]$,  for all $ x \in \Theta_\Cb$,  where the lower and upper limits are given by \cite{de1994probability}
 \begin{eqnarray}
 \underline{p}^{\Db\downarrow \Cb}_x & = & \max \left \{ \sum_{y:y\downarrow x} \underline{p}^\Db_y,\;\;1-\sum_{y:y\downarrow \Theta_\Cb\setminus \{x\}} \overline{p}^\Db_y \right\}, \\
  \overline{p}^{\Db\downarrow \Cb}_x & =  & \min\left \{ \sum_{y:y\downarrow x} \overline{p}^\Db_y,\;\; 1 - \sum_{y:y\downarrow \Theta_\Cb\setminus \{x\}} \underline{p}^\Db_y \right\},
  \end{eqnarray}
respectively.  Marginalisation is the inverse operation of the vacuous extension, that is, $(K^{\Cb\uparrow \Db})^{\downarrow \Cb} = K^\Cb$. However, in general, the vacuous extension is not the inverse of marginalisation.

\subsubsection{Axioms}
Assuming the set of valuations $\Phi_\Db$ consist of credal sets defined by coherent probability intervals on singletons, we want to verify that the axioms of a valuation algebra hold. Staring with (A1), we verify that $(\Phi_\Db,\otimes)$, is a commutative monoid. First, the set  $\Phi_\Db$ is closed under  combination  (\ref{e:comb_op}), because of  (\ref{e:cs3}).  Next,  it is straightforward to verify that both commutativity
\[ K_1^\Db \otimes K_2^\Db  = K_2^\Db \otimes K_1^\Db \]
and associativity
\[ K_1^\Db \otimes (K_2^\Db \otimes K_3^\Db) = (K_1^\Db \otimes K_2^\Db) \otimes K_3^\Db\]
hold because  multiplication, addition, $\min$ and $\max$ operations, which feature in (\ref{e:combP_lo}) and (\ref{e:combP_up1}) are commutative and associative. Next we  find the identity valuation $K_e^\Db \in \Phi_{\Db}$, such that $K_e^\Db \otimes K^\Db = K^\Db$ for all $K^\Db\in\Phi_\Db$. It turns out that $K^D_e$ contains only one PMF: the uniform PMF on $\Theta_D$, the same identity element as in the valuation algebra of PMFs. It is interesting to note that the vacuous credal  set $\mathcal{P}^\Db$ is the absorbing element of $\Phi_\Db$, that is  $K^\Db \otimes \mathcal{P}^\Db = \mathcal{P}^D$ for every $K^\Db\in\Phi_\Db$. The axiom of labeling (A2) follows from the way we combine valuations on different domain via vacuous extension. Marginalization axiom (A3) follows directly from its definitions. Axioms (A4) and (A5) follow from the definitions of the combination and marginalisation as convex sets of PMFs.  Axiom (A6) follows directly from the definition of identity element and the definition of combination over different domains.

Next we discuss a practical implementation of a credal valuation network using local computation.

\section{Credal Valuation Network}

Valuation network computes the marginal $(\otimes \Phi)^{\downarrow\Db_o}$ on local domains, that is, without explicitly computing the joint valuation on the full domain $\Vb$.
This computation is carried out using  the {\em fusion algorithm}, which eliminates sequentially all variables $X \in \Vb\setminus \Db_o$ which are of no interest to the inference problem \cite{shenoy_92, Haenni2004, Benavoli2009}. The fusion algorithm is applied over a structure called the binary joint tree (BJT), where all
combinations  are carried on pairs of valuations, that is on a binary basis (two-by-two). Finally, marginals are computed by means of a message-passing scheme among the nodes of the BJT. Full details of software implementation of a generic valuation network can be found in \cite{shenoy_92, Haenni2004, Benavoli2009}, and therefore will not be repeated here. Instead, we will focus on  the computation of combination operator (\ref{e:cs2}). Implementation of the vacuous extension and marginalisation is rather straightforward.

\subsection{Implementation of the combination operator}

The combination operator of (\ref{e:comb_op}) results in the credal set specified  with interval probabilities on singletons, given by (\ref{e:cs3}).  The key is to compute the lower and upper probabilities of these intervals, given by (\ref{e:combP_lo}) and (\ref{e:combP_up}). We can reformulate the optimisation problem in (\ref{e:combP_lo}) by introducing a scalar variable $\nu$ as follows \cite{ristic2020tutorial}:
\begin{equation}
    \underline{p}_i =  \max \nu, ~~	s.t.~~
	\min_{\substack{p_1\in K_1^D\\p_2 \in K_2^D}}~~ \sum_{x\in\Theta_\Db} \left(\mathbf{1}_{\{x_i\}}(x) -\nu\right) p_1(x)p_2(x)\geq 0,
	\label{e:opt1}
\end{equation}
where $\mathbf{1}_{\{x_i\}}(x)$ is the indicator function (it equals $1$ if $x=x_i$ and zero otherwise). Note that (\ref{e:opt1}) involves two optimisation problems, a minimisation and a maximisation.

The minimisation problem in (\ref{e:opt1}) can be written in a vector from. First, recall that both $K_1^\Db$ and $K_2^\Db$ are specified in the form of probability intervals, cf. (\ref{e:cs2}). Let us denote the lower probability envelope of $K_m^\Db$, for $m=1,2$ with $\underline{p}_{mi}$, for $i=1,\dots,n$, where $n$ is the cardinality of $\Theta_\Db$. Accordingly, the upper probability envelope of $K_m^\Db$, for $m=1,2$, is specified with $\overline{p}_{mi}$, for $i=1,\dots,n$. Minimisation in (\ref{e:opt1}) can be written as
\begin{align}
    \min_{\pb_1,\pb_2} & ~~~ \pb_1^T \, \mbox{diag}[\cb_i]\, \pb_2 \label{e:bilin} \\  \text{ subject to }  & \Ab\pb_1 \leq \bb_1, \;
    \Ab\pb_2 \leq \bb_2, \nonumber \\
    &\pb_1 \geq 0, \text{ and } \pb_2 \geq 0, \nonumber
\end{align}
where  $\pb_m$, for $m=1,2$, is the probability vector given by $\pb_m = [p_m(x_1), \, p_m(x_2),\, \dots, p_m(x_n)]^T$.  Notation $\mbox{diag}[\cb_i]$ denotes an $n\times n$ diagonal matrix with vector $\cb_i$ along the diagonal. The $i$th element of vector $\cb_i$ is $1-\nu$, while all other elements are equal to $-\nu$.  Dimension of matrix $\Ab$ is $(2n+2) \times n$ and is given by
\begin{equation}
    \Ab = \left [ \begin{matrix} -1 & 0 & 0 & \cdots & 0\\
                                1 & 0 & 0 & \cdots  & 0\\
                                0 & -1 & 0 & \cdots & 0\\
                                0 & 1  & 0 & \cdots & 0\\
                                \vdots & \vdots & \vdots & \ddots & \vdots \\
                                0 & 0 & 0 & \dots & -1 \\
                                0 & 0 & 0 & \dots & 1\\
                                1 & 1 & 1 & \dots & 1\\
                                -1 & -1 & -1 & \dots & -1\end{matrix}
                                \right]
\end{equation}
Finally vector $\bb_m$ for $m=1,2$ of dimension $2n+2$ includes the input probability limits of $K_m^\Db$, and is given by:
\begin{equation}
    \bb_m = \left[ \begin{matrix}  -\underline{p}_{m1}& \overline{p}_{m1} & -\underline{p}_{m2}& \overline{p}_{m2} & \cdots &-\underline{p}_{mn}& \overline{p}_{mn} & 1 & -1 \end{matrix}\right]^T.
\end{equation}
Given $\nu$, the minimisation problem in (\ref{e:bilin})
is {\em bilinear} in the unknowns $\pb_1$ and $\pb_2$.
If for example $p_1$ is precise, then the minimisation problem becomes linear in the unknown $p_2$ (and vice versa) and can efficiently be solved by linear programming.
Maximisation over $\nu$ in (\ref{e:opt1}) can be solved by a bisection method.

The solution for upper probability  (\ref{e:combP_up}), following the same approach, can be reformulated as follows:
\begin{equation}
    \overline{p}_i =  1 - \max \nu, ~~	s.t.~~
	\min_{\substack{p_1\in K_1^D\\p_2 \in K_2^D}}~~ \sum_{x\in\Theta_\Db} (1 - \mathbf{1}_{\{x_i\}}(x) -\nu) p_1(x)p_2(x)\geq 0.
	\label{e:opt2}
\end{equation}
Minimisation in (\ref{e:opt2}) can also be  written in the vector form (\ref{e:bilin}). The only difference is in the specification of vector $\cb_i$: its $i$th element is now $-\nu$, while all other elements are equal to $1-\nu$. The bilinear optimisation problem (\ref{e:bilin}) is solved using the Gurobi software for optimisation \cite{gurobi}. Further details are given in Appendix B.


\subsection{Demonstration and comparison}
\label{s:demo}

In this section we solve the ``Arrival delay'' problem introduced in Sec. \ref{s:valuations} using a credal valuation network (CVN) developed for this problem, based on the theoretical foundations described above.  Subsequently we compare its solution to the solution obtained using the corresponding evidential network (EN), described in \cite{Benavoli2009,benavoli_ristic_2013 }. For comparison of the two reasoning solutions we adopt the framework for assessment proposed in \cite{ristic2021performance}. The main premise of this framework is that the system under investigation (in our case, the arrival delay) is uncertain only due to stochastic variability (that is, all probabilistic models in reality are precise). However, these precise probabilistic models are only partially known by the the systems for reasoning,  the CVN and the EN.  The  solutions obtained using the CVN and the EN are therefore evaluated against the true solution, obtained using the valuation network of (precise) PMFs.

\begin{table} [tbh]
\caption{\em Variables of the valuation network in Fig. \ref{f:va}}
\begin{center}
\begin{tabular}{lll}
\hline Variable & Name & Frame (in days) \\
\hline
 A & Arrival delay & $\Theta_A=\{0,1,\dots,4\}$ \\
 D & Departure delay & $\Theta_D = \{0,1,2\}$\\
 T & Travel delay & $\Theta_T=\{0,1,2\}$ \\
 L & Loading delay  & $\Theta_L=\{0,1\}$ \\
 S & Service delay & $\Theta_S=\{0,1\}$ \\
 W & Weather delay & $\Theta_W=\{0,1\}$ \\
 R & Repair on sea & $\Theta_R=\{0,1\}$ \\
 \hline
\end {tabular}
\end{center}
\label{t:var}
\end{table}

The list of variables and their frames for the ``Arrival delay'' valuation network (shown in Fig. \ref{f:va}) is summarised in Table \ref{t:var}. Each frame represents a set of the integers corresponding to the number of days. For example, delays due to loading, service, weather or repair, can be at most 1 day. The valuations, expressing the relationships between the variables,  are specified in Table \ref{t:val}. The first row of Table \ref{t:val} states that the arrival delay A is a superposition of D and T (with equal weights), expressed as A = D + T, and this relationship is true with probability $1.0$.  The reasoning systems, however, can only assume that this relationship is true with a probability in the interval $[0.96,1.00]$, and therefore need to deal with additional epistemic uncertainty. According to row 2 of Table \ref{t:val}, the relationship  D~=~L~+~S is in reality true with probability $0.91$ (i.e. other causes can be involved). The reasoning systems, on the other hand, can only have confidence in this relationship in the interval $[0.90,0.92]$. Valuation $\varphi_4$ is specified by an implication rule, which is true with probability $0.89$. Both CVN and EN only know that this probability is in the interval $[0.88,0.91]$. Note that $\varphi_5$, $\varphi_6$ and $\varphi_7$ are  expressions of uncertain information about a single variable, i.e. about L, S and W, respectively. Valuations $\varphi_1$, $\varphi_2$, $\varphi_3$ and $\varphi_4$ can be considered as  {\em domain knowledge},  while $\varphi_5$, $\varphi_6$ and $\varphi_7$ are the pieces of information received possibly a few days before the departure of the ship.

\begin{table} [bht]
\caption{\em Valuations of the  network in Fig. \ref{f:va}}
\begin{center}
\begin{tabular}{ccccc}
\hline Valuation & Domain & Knowledge & True probability & Interval probability \\
\hline
$\varphi_1$ & $\{\text{A,D,T}\}$ & A = D + T & $1.0$            & $[0.96, 1.00]$\\
$\varphi_2$ & $\{\text{D,L,S}\}$ & D = L + S & $0.91$           & $[0.90, 0.92]$\\
$\varphi_3$ & $\{\text{T,R,W}\}$ & T = R + W & $0.94$           & $[0.92, 0.95]$\\
$\varphi_4$ & $\{\text{S,R}\}$ &  If S $=1$ then R $=0$ & $0.89$ & $[0.88,0.91]$\\
$\varphi_5$ & $\{ \text{L}\}$ & L $=1$ & $0.82$                 & $[0.80,0.83]$\\
$\varphi_6$ & $\{ \text{S}\}$ & S $=0$ & $0.73$                 & $[0.71,0.74]$\\
$\varphi_7$ & $\{ \text{W}\}$ & W $=1$ & $0.64$                 & $[0.62,0.65]$\\
\hline\end{tabular}
\end{center}
\label{t:val}
\end{table}

\begin{figure}[htb]
\centerline{\includegraphics[width=0.49\linewidth]{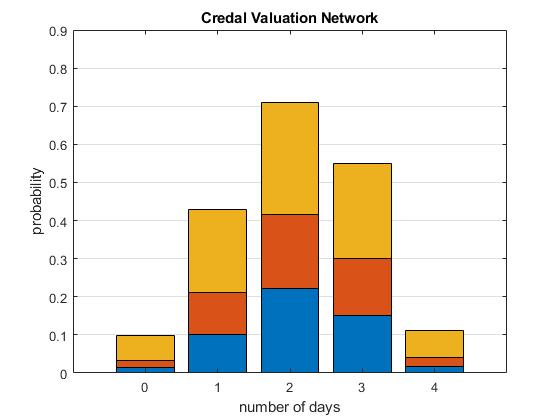}\hspace{1mm} \includegraphics[width=0.49\linewidth]{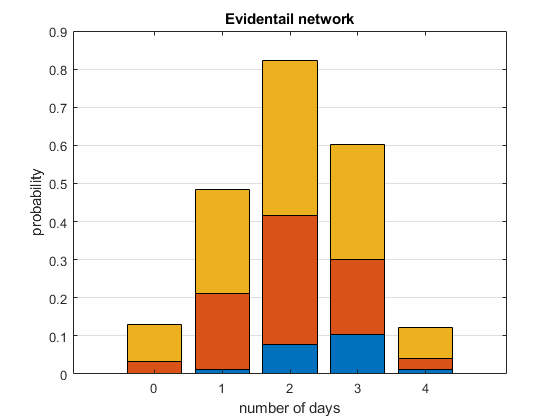}}
\centerline{\footnotesize (a)\hspace{8cm}(b)}
 \caption{\footnotesize A bar graph representation of numerical results in Table \ref{t:res}: the marginal probability distribution of  variable A , using (a) the CVN and (b) the EN. The height of the orange bar equals the ``ground thruth''.}
 \label{f:bar}
\end{figure}

The output of a valuation network in this example is the joint valuation $\varphi_1\otimes\cdots\otimes\varphi_7$, marginalised to variable A. This marginal probability distribution is presented  in Table \ref{t:res}, for
three valuation networks. The valuation network of  PMFs (VN-PMF), see Sec. \ref{s:VA_PMF}, uses the true precise  probabilities (column 4 in Table \ref{t:val}) assigned to available knowledge for inference (column 3 in Table \ref{t:val}). Its output (the second column in Table \ref{t:res}) is the precise marginal distribution of variable A, and is considered the ``ground truth'' in this example. The results obtained using the CVN and the EN are presented in rows 3 and 4, respectively, of Table \ref{t:res}. The same results are displayed as two bar graphs in Fig. \ref{f:bar}. We point out that in this example we were able to use an exact method for transforming the
interval probabilities (given in the fifth column of Table \ref{t:val}) to belief functions\footnote{Note that the conditions of Proposition 14 of \cite{de1994probability} are satisfied in our example. Then, the belief function  corresponds to  lower probabilities of the entire power set, using formulae in Sec. 4.4 of \cite{Augustin2014}. } meaning that the input for both the CVN and the EN is  identically uncertain information. The output of the EN is the belief-plausibility pair on the elements of $\Theta_A$.

\begin{table} [bht]
\caption{\em Marginal probability distribution of variable A}
\begin{center}
\begin{tabular}{c|ccc}
\hline A (days)  & VN-PMF & CVN & EN  \\
\hline
$0$ & $0.034$   & $[0.015,  0.099]$ &  $[0.000,   0.129]$\\
$1$ & $0.210$   & $[0.101,  0.428]$ &  $[0.012,   0.485]$\\
$2$ & $0.415$   & $[0.221,  0.711]$ &  $[0.076,   0.823]$\\
$3$ & $0.301$   & $[0.151,  0.549]$ &  $[0.105,   0.603]$\\
$4$ & $0.040$   & $[0.016,  0.111]$ &  $[0.011,   0.121]$ \\
\hline\end{tabular}
\end{center}
\label{t:res}
\end{table}

Note from Table \ref{t:res} and Fig. \ref{f:bar} that both
the CVN and the EN express the marginal probability of variable A with probability intervals. Importantly, these intervals always contain the ``ground truth'' probability, obtained using the VN-PMF. For example, according to row 1 of Table \ref{t:res}, the probability that arrival delay is 0 days is 0.034, and this value is contained in both interval $[0.015,  0.099]$ for CVN and $[0.000,   0.129]$ for EN. However, observe that the intervals are much tighter, and therefore, the epistemic uncertainty smaller, using the CVN, rather than the EN,  for inference. In order to quantify performance, we can apply the evaluation method proposed in \cite{ristic2021performance} to quantify the accuracy of the CVN and EN. This method computes the distance between the ``ground truth'' PMF $[p_1,\dots,p_n]$ and the credal set, expressed with the lower probability envelope $[\underline{p}_1,\dots,\underline{p}_n]$ and the upper probability envelope  $[\overline{p}_1,\dots,\overline{p}_n]$,  as follows:
\begin{equation}
    D = \left[1+\exp\left\{ -\frac{1}{n} \sum_{i=1}^n \log \frac{(\overline{p}_i - \underline{p}_i)}{Pr\{p_i \in [p_i]\}}  \right\}\right]^{-1}. \label{e:D}
\end{equation}
The distance $D$ takes values from interval $[0,1]$, with smaller values indicating a smaller distance.  The values of distance $D$ for the output of the CVN and the output of the EN (according to Table \ref{t:res}), are  computed as $0.18$ and $0.23$, respectively. Hence, we conclude that in terms of the accuracy of reasoning, the CVN outperforms the EN.

All three aforementioned valuation networks were implemented in MATLAB (Gurobi optimisation software can be called from MATLAB) and applied using the same sequence of elimination variables. The measured computation time of the VN-PMF, the CVN and the EN on this example is $0.08$, $16.12$ and $0.48$ seconds, respectively. This example involves a small number of focal sets in the  EN and hence the EN is faster to run than the CVN.

\section{Conclusions}

The paper presented the theoretical foundations and discussed a practical implementation of a valuation  network  for  reasoning,  where uncertain pieces of collected information and domain knowledge are  expressed  as credal  sets  defined  by  the  coherent  probability intervals. This framework was adopted as a generalisation of a valuation network of probability mass functions, for situations where both the aleatory and epistemic uncertainties are present in the knowledge-base and observations. The developed credal valuation network was demonstrated on a small scale example and compared to the corresponding evidential network (which represents valuations using the Demspter-Shafer belief functions). The result of reasoning using the CVN is less uncertain and therefore more desirable than the result obtained using the evidential network. The future work will focus on improving the computational efficiency of the CVN and its application  to realistic problems in military surveillance.

\appendix
\centerline{}
\noindent{\bf A. Conditions of coherence for upper and lower probabilities}\\
Condition (\ref{e:cond1}) is simple to verify. Consider a PMF $p\in K^X$, where $K^X$ is a credal set specified by interval probabilities as in (\ref{e:cs2}). Then we can write:
    $\underline{p}_i \leq p(x_i) \leq \overline{p}_i$,
for $i=1,\dots,|\Theta_X|$. If we perform summation over index $i$, then we have:
$\sum_{i}\underline{p}_i \leq \sum_i p(x_i) \leq \sum_i \overline{p}_i.$
Since $p$ is a PMF, then the middle term $\sum_i p(x_i) =1$ and condition (\ref{e:cond1}) immediately follows.

Next we show that condition
\begin{equation} \sum_{j\neq i} \underline{p}_j + \overline{p}_i \leq 1
\label{e:app0}
\end{equation}
is obtained from the statement that there exist a PMF $p^i\in K^X$ such that it {\em reaches} the  upper probability $\overline{p}_i$, that is $p^i(x_i)=\overline{p}_i$. From the definition of credal set $K^X$ (\ref{e:cs2}), for every $j=1,\dots,|\Theta_X|$ we have $ \underline{p}_j \leq p^i(x_j).$ If we perform summation of both sides of this inequality over index $j$, such that $j\neq i$, we obtain:
\begin{equation}
    \sum_{j\neq i} \underline{p}_j \leq \sum_{j\neq i} p^i(x_j)
    \label{e:app1}
\end{equation}
Adding the term $\overline{p}_i$ to both sides of (\ref{e:app1}), we have
\begin{equation}
    \sum_{j\neq i} \underline{p}_j + \overline{p}_i\leq \sum_{j\neq i} p^i(x_j) + \overline{p}_i.
    \label{e:app2}
\end{equation}
 Since  $\overline{p}_i= p^i(x_i)$ and $p^i$ is a PMF, the sum on the right hand side of (\ref{e:app2}) equals $1$, which proves (\ref{e:app0}).

\centerline{}
\noindent{\bf B. Solving bilinear optimisation using Gurobi}\\
In order to apply Gurobi \cite{gurobi}, we need to express the optimisation problem (\ref{e:bilin}) in the form:
\begin{align}
    \min_{\xb}& ~~~ \gb^T\,\xb  \label{e:bilin1} \\  \text{ subject to }  & \Ab^*\xb \leq \bb^*, \nonumber \\
    \text{ and } & \xb^T \Qb \, \xb + \xb^T \qb \leq r.\nonumber
    \label{e:bilin1}
\end{align}
The first constraint above  is linear, while the second is quadratic.
Note that we can rewrite the objective of optimisation in  (\ref{e:bilin}), that is,  $\min
~ \pb_1^T \, \mbox{diag}[\cb_i]\, \pb_2$, as follows:
\begin{equation} \min a  \text{ s.t. } \pb_1^T \, \mbox{diag}[\cb_i]\, \pb_2 \leq a.
\label{e:constA}
\end{equation}
The constraint in (\ref{e:constA}) will be expressed as a quadratic constraint in (\ref{e:bilin1}). Minimisation (\ref{e:bilin}) can now be written in the form of (\ref{e:bilin1}) with the following definitions:
\begin{align*}
\xb & = \left[\begin{matrix} a & \pb_1^T & \pb_2^T\end{matrix}\right]^T\\
\gb & = \left[ \begin{matrix} 1 & 0 & \cdots & 0\end{matrix}\right]^T\\
\Ab^* & = \left [\begin{matrix}  \mathbf{0}_{2(n+1)\times 1} & \Ab & \mathbf{0}_{2(n+1)\times n} \\ \mathbf{0}_{2(n+1)\times 1} &  \mathbf{0}_{2(n+1)\times n} & \Ab
\end{matrix}\right]\\
\bb^* & = \left [ \begin{matrix}  0 & \bb_1^T & \bb_2^T\end{matrix}\right]^T\\
\Qb & = \left[\begin{array}{cc}
 \mathbf{0}_{(2n+1)\times (n+1)} & \begin{matrix} \mathbf{0}_{1\times n}\\  \mbox{diag}[\cb_i] \\  \mathbf{0}_{n\times n} \end{matrix} \end{array}\right]\\
 \qb & = - \gb\\
 r & = 0,
\end{align*}
where $\mathbf{0}_{a\times b}$ is a zero matrix of dimension $a\times b$. Optimisation is non-convex.

\bibliographystyle{IEEEtran}
\bibliography{Reas,IP_ref}

\end{document}